\documentclass[10pt,twocolumn,letterpaper]{article}

\usepackage{wacv}
\usepackage{times}
\usepackage{epsfig}
\usepackage{graphicx}
\usepackage{amsmath}
\usepackage{amssymb}
\usepackage{pifont}
\usepackage{mwe}
\usepackage{multirow}
\usepackage{tabularx}
\usepackage{subfig}
\usepackage{tablefootnote}

\newcommand{\fref}[1]{Figure~\ref{#1}}
\newcommand{\tref}[1]{Table~\ref{#1}}
\newcommand{\eref}[1]{Equation~\ref{#1}}
\newcommand{\sref}[1]{Section~\ref{#1}}
\wacvfinalcopy % *** Uncomment this line for the final submission

\ifwacvfinal
\def\assignedStartPage{1} % *** Enter the assigned starting page number (instead of 9876)
\fi

\ifwacvfinal
\usepackage{hyperref}
\hypersetup{breaklinks=true,bookmarks=false}
\else
\usepackage[pagebackref=true,breaklinks=true,colorlinks,bookmarks=false]{hyperref}
\fi

\ifwacvfinal
\else
\fi

\begin{document}

%%%%%%%%% TITLE
\title{Self-Supervised Shadow Removal}

\author{Florin-Alexandru Vasluianu\\
ETH Zurich\\
{\tt\small fvasluianu@student.ethz.ch}
\and
Andrés Romero\\
 ETH Zurich\\
{\tt\small  roandres@ethz.ch}
\and
Luc Van Gool\\
 ETH Zurich\\
{\tt\small  vangool@vision.ee.ethz.ch}
\and
Radu Timofte \\
 ETH Zurich\\
{\tt\small  radu.timofte@vision.ee.ethz.ch}
}
\maketitle
%\thispagestyle{empty}

%%%%%%%%% ABSTRACT
\begin{abstract}
Shadow removal is an important computer vision task aiming at the detection and successful removal of the shadow produced by an occluded light source and a photo-realistic restoration of the image contents.
Decades of research produced a multitude of hand-crafted restoration techniques and, more recently, learned solutions from shadowed and shadow-free training image pairs.
In this work, we propose an unsupervised single image shadow removal solution via self-supervised learning by using a conditioned mask. In contrast to existing literature, we do not require paired shadowed and shadow-free images, instead we rely on self-supervision and jointly learn deep models to remove and add shadows to images.
We validate our approach on the recently introduced ISTD and USR datasets. We largely improve quantitatively and qualitatively over the compared methods and set a new state-of-the-art performance in single image shadow removal.

\end{abstract}

%%%%%%%%% BODY TEXT
%%%%%%%%%%%%%%%%%%%%%%%INTRODUCTION%%%%%%%%%%%%%%%%%%%%%%%%%

\section{Introduction}
\label{sec:introduction}

In an image, a shadow~\cite{stamminger2002perspective} is the direct effect of the occlusion of a light source.
By inducing a steep variation in an image region, the shadow impacts the performance of other vision tasks such as image segmentation~\cite{felzenszwalb2004efficient,achanta2012slic},
semantic segmentation~\cite{shelhamer2016fully,garcia2018survey}, object recognition~\cite{uijlings2013selective,arbelaez2014multiscale,he2017mask} or tracking~\cite{kaewtrakulpong2002improved,kristan2015visual,danelljan2014adaptive}. 

In contrast to the unshadowed pixels, the shadow alters the observation of the scene contents by a combination of degradations in illumination, color, detail, and noise levels.
The shadow removal task is, essentially, an image restoration task aiming at recovering the underlying content.
Many methods~\cite{sanin2012shadow,nguyen2017shadow} have been proposed for detecting and removing shadows from images.% \AR{you could reference some more for "removing"}

The introduction of large-scale datasets of shadowed and shadow-free image pairs such as SRD~\cite{SRDDESHADOW}, ISTD~\cite{ISTDwang2018STCGAN} or USR~\cite{USRhu2019mask} allowed the formulation of the shadow removal process as a regression problem.
One of the major challenges is to learn a physically plausible transformation, regardless of the semantic or illumination inconsistencies that may be encountered in the data. Thanks to the advent of Generative Adversarial Networks~\cite{goodfellow2014generative} and its flexibility learning complex distributions, recent efforts ~\cite{USRhu2019mask,Zhang2020RISGANER} have successfully modeled the shadow removal task as an Image-to-Image translation problem~\cite{isola2017image}. 
However, it has been found ~\cite{Odena2016DeconvolutionAC} that the learned shadow removal transformations are highly prone to artifacts produced in the downsampling/upsampling phases of the translation encoder/decoder model, and moreover, the tendency of the deshadowed image regions to be blurry~\cite{USRhu2019mask,zhang2016colorful}. In order to circumvent these problems, recent solutions~\cite{ISTDwang2018STCGAN,USRhu2019mask,Le_2019_ICCV} have proposed carefully designed robust loss functions, producing thus high quality photo-realistic deshadowed results with low pixel-wise restoration errors.

As the shadow removal is a perceptual transformation, the usage of a perceptual score based on learnt features, on different levels of complexity \cite{zhang2018perceptual}, enabled the exploitation of some invariants over the shadow removal or addition transformation. The increased amount of information used in training is expected to induce additional degrees of control such that the learning procedure can be faster, and the results produced will be better, both in terms of fidelity metrics and perceptual scores. 

In this work, we propose an unsupervised single image shadow removal solution via self-supervised learning.
Our method exploits several observations made on the shadow formation process and employs the cyclic consistency and the Generative Adversarial Nets (GANs)~\cite{goodfellow2014generative} paradigms as inspired by the CycleGAN~\cite{CycleGAN2017}, a seminal architecture for learning image-to-image translation between two image domains without paired data.

\begin{figure}[t] 
    \centering
    \subfloat[Forward step]{%
            \includegraphics[width=0.9\linewidth]{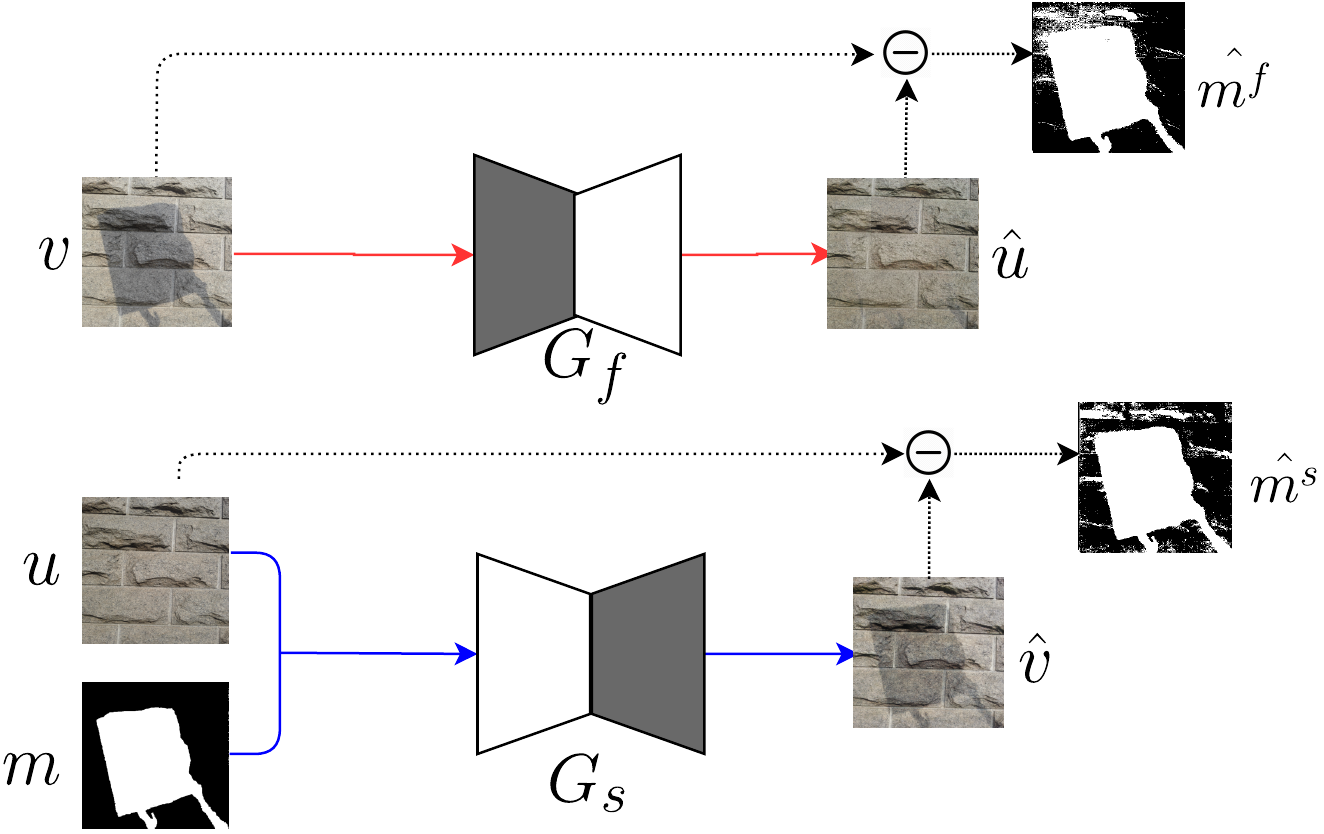}
            \label{fig:forward_step}
        }
    \vfill
    
    \subfloat[Reconstruction step]{%
            \includegraphics[width=0.9\linewidth]{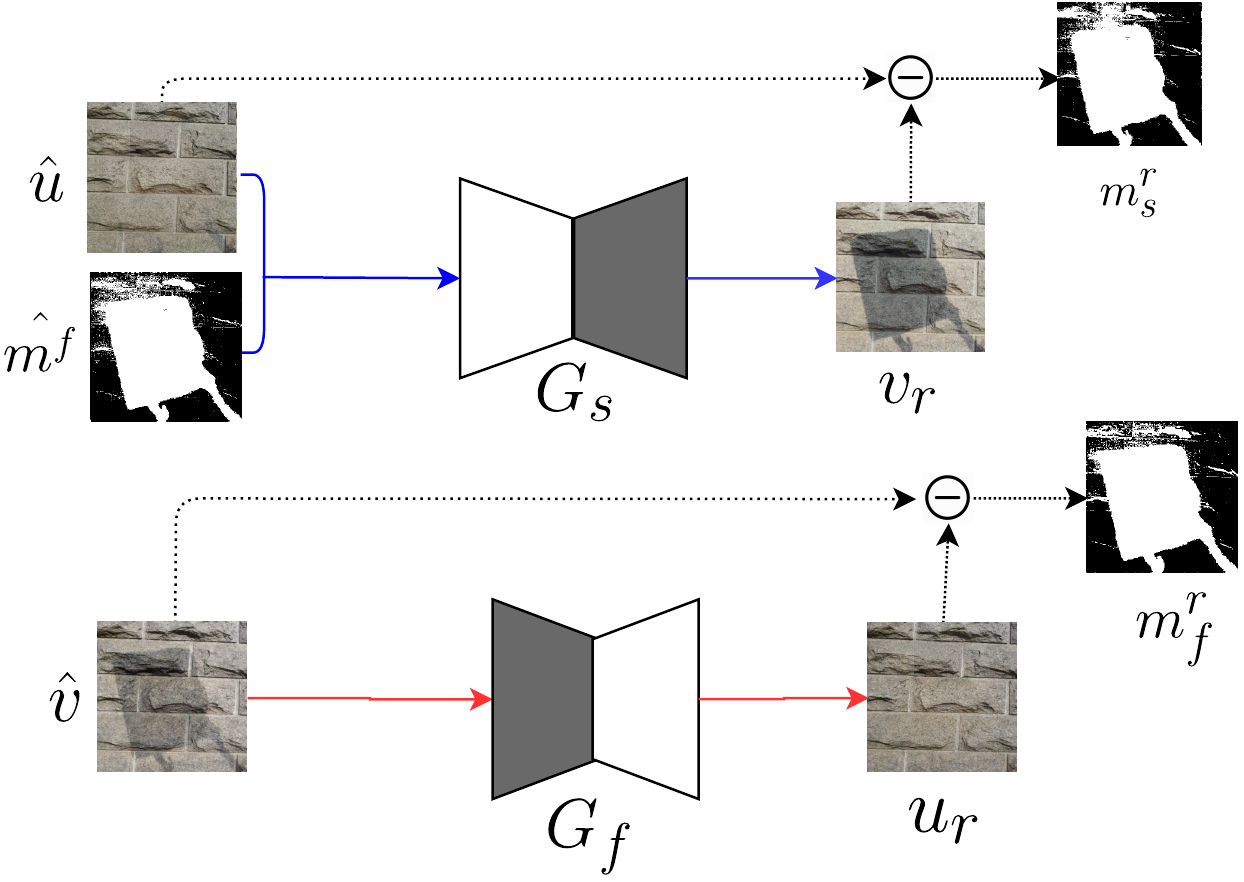}
            \label{fig:backward_step}
        }%
    \caption{The forward (top) and the reconstruction step (bottom). As a convention, red lines were used for the manipulation involving shadow affected input, blue lines for the shadow free input, and the black lines for the mask computation operations. \textit{u} and \textit{v} could or could not be paired data. Our training framework uses \textit{v}'s mask (\textit{m}) to insert it in $G_s$. In a paired setting $\hat{v}$ should resemble $v$, while not the case for an unpaired one.}
    \label{fig:overview}
\end{figure}

An overview of our method is depicted in \fref{fig:overview}. By assuming that each dataset contains shadow images with respective masks, either on paired or unpaired settings, the core of our method is to exploit the given mask information in a self-supervised fashion by inserting randomly created shadow masks into the training framework (input \textit{m} during the forward step in \fref{fig:overview}.a), and reconstructing the original input using cycle-consistency (\fref{fig:overview}.b).

A critical observation is that we do not need to impose strong pixel-wise fidelity losses in our solution, but rather capture contents and general texture and colors, which are inherently perceptual. 
This helps our model to produce higher quality results, in a smaller number of training epochs with respect to the literature.

%%%%%%%%%%%%%%%%%%%%%%%%%%%%%%%%%%%%%%%%%%%%%%%%%%%%%%%%%%%%%

%%%%%%%%%%%%%%%%%%%%RELATED WORK%%%%%%%%%%%%%%%%%%%%%%%%%%%%

\section{Related Work}
\label{sec:related_work}

Shadow removal is not a recent problem in the computer vision community, and despite recent efforts in deep learning and generative modeling it is still a challenging problem. 

Early methods tackled this problem by using the underlying physical properties of the shadow formation. They were based on image decomposition as a combination of shadow and shadow-free layers~\cite{finlayson2002removing,Finlayson_entropyminimization}, or on an early shadow detection followed by a color transfer from the shadow-free region to the shadow affected region in the local neighborhood~\cite{10.1145/1243980.1243982,Shor:2008:TSM,7893803}. 

The variety of shadow generation systems (\eg shapes, size, scale, illumination, etc) implies an increased complexity in shadow model parameters computation, and consequently, models parameterized with these properties are known for not being able to handle shadow removal in complex situations~\cite{6909646}. A step forward in this direction is a two-stage model for shadow detection and removal, respectively, thus increasing the generalization performance. In order to successfully detect the shadow, earlier works~\cite{guo2013paired,gong2014interactive} proposed to include hand-crafted features such as image intensity, texture, or gradients.

The evolution of the Convolutional Neural Networks (CNNs) enabled the propagation of these learnable features along with the layers of the model, and \cite{6909646,articleKhan} proposed solutions using CNNs for shadow detection and a Bayesian model for shadow removal. Moreover, \cite{gong2014interactive} pioneered using an unsupervised end-to-end auto-encoder model to learn a cross-domain mapping between shadow and shadow-free images. %...\AR{Could you mention some shortcoming of this method here?}.
However, the need for manual labeling in the pre-processing step in order to produce an estimation over the shadow mask limits this method, both in terms of the complexity of the addressed light-occluder systems and the bias injection.

Qu~\etal~\cite{SRDDESHADOW} proposed a model based on three networks extracting relevant features from multiple views and aggregating them to recover the shadow-free image. The G-net (Global localization network) will extract a high-level representation of the scene, the A-net (Appearance modeling network), and the S-net (Semantic modeling network) will use this representation to produce a shadow matte used in the shadow removal task.

The importance of the localization information is acknowledged by Hu~\etal~\cite{hu2019direction}, where the shadows were detected and removed using the idea of a Spatial Recurrent Neural Network~\cite{bell2016inside} by exploring the direction-aware context.

Subsequently, since the introduction of Generative Adversarial Networks (GANs)~\cite{goodfellow2014generative}, the dominant strategy is to learn an image-to-image mapping function using an encoder/decoder architecture. 
The de-facto methodologies for image-to-image translation for paired and unpaired data are pix2pix~\cite{isola2017image} and CycleGAN~\cite{CycleGAN2017}, respectively. In the former, it is assumed a single transformation between shadow and deshadowed regions, while in the latter, in order to deal with the unsupervised nature of the data, there are two different models for shadow removal and addition, and a cycle-constraint loss ensures the flow from one domain to another.

Following this trend, \cite{ISTDwang2018STCGAN} proposed a model based on Conditional GANs~\cite{Mirza2014ConditionalGA} using paired data, where they deployed 2 stacked conditional GANs aiming shadow detection and then, with the information computed, shadow removal.

Recently, Le~\etal~\cite{Le_2019_ICCV} proposed a model based on two neural networks able to learn the shadow model parameters and the shadow matte. The main limitation of the model is the usage of a simple linear relation as the light model. However, by using the same occluder there can be multiple light sources to produce non-homogeneous shadow areas that can not be described by a linear model, and therefore, the performance of the model is expected to drop. Nonetheless, if the assumption made about the uniqueness of the light source holds, the method is able to produce realistic results.

Despite the recent effort in shadow removal literature, all prior methods rely on the assumption of paired datasets for shadow and shadow-free images. Going in an unsupervised direction, \cite{USRhu2019mask} developed MaskShadowGAN using a vanilla CycleGAN~\cite{CycleGAN2017} approach, where the shadow masks are computed as a binarization of the difference, by thresholding it using Otsu's algorithm. In contrast, we formulate a component in the training objective that is going to offer some bounds in the evolution of the synthetically-generated shadow masks (used as input in the reconstruction), which will increase the degree of control in the training procedure, improving the performance both in fidelity and perceptual metrics. Moreover, the decrease of the loss in the training/validation procedure will be more pronounced, even if the generators we used are characterized by a smaller number of parameters. 
%%%%%%%%%%%%%%%%%%%%%%%%%%%%%%%%%%%%%%%%%%%%%%%%%%%%%%%%%%%%

%%%%%%%%%%%%%%%%%%%APPROACH%%%%%%%%%%%%%%%%%%%%%%%%%%%%%%%%%

\section{Proposed Method}
\label{sec:proposed_method}

\subsection{Problem Formulation} 
Considering the shadow image domain $X$ and the shadow-free image domain set $Y$, we are mainly interested to learn the mapping function $G_{f}: X \rightarrow Y$. Existing techniques~\cite{isola2017image} rely on a critical dataset assumption of having access to paired images, \ie the same scene with and without shadows. As we will show in \sref{sec:experimental_results}, this assumption does not always hold, and having an unsupervised approach leads, surprisingly, to better performance. To this end, we assume a subset of unpaired images $T = \{(u,v) | u \in Y, v \in X\}$.

\subsection{Overall scheme}

The overall scheme of our method is presented in \fref{fig:overview}. Our system is based on the vanilla CycleGAN approach~\cite{CycleGAN2017} with two generators and two discriminators for each domain: $G_s$ and $G_f$, being shadow addition and shadow removal, respectively. In detail, the shadow addition network receives two inputs: an image and a binary mask depicting the shadow. The shadow removal network only receives an image. Formally, $\hat{u}=G_{f}(v)$ and $\hat{v}=G_{s}(u,m)$, for removal and addition respectively (\fref{fig:forward_step}), being $m$ a randomly generated mask. To close the cycle-consistency loop, we use self-supervision to reconstruct the original inputs (\fref{fig:backward_step}). We extensively explain this process in~\sref{sec:s3}.

Besides our self-supervised training framework, an important ingredient of our method relies on the carefully designed loss functions we proceed to explain in the following section.\footnote{Codes will be made publicly available at\\ \url{https://github.com/fvasluianu-cvl/SSSR.git}}

\subsection{Objectives and losses}
For simplicity and sake of clarity, it is important to mention that we define our losses regardless of the training settings (paired or unpaired) and the transformation mapping (inserting or removing shadow), and instead we use placeholders. We will make a clear distinction at the end of this section.

\subsubsection{Pixel-wise losses}
Our main motivation to build an unsupervised shadow removal comes from an observation we illustrate in~\fref{fig:ISTD_differences}. On the one hand, there are pixel-wise inconsistencies (\eg, different lighting conditions outside the shadow or content misalignment) for paired images in the ISTD dataset~\cite{ISTDwang2018STCGAN}, so building a model under this assumption compromises the performance. On the other hand, using loss functions based solely on a pixel-wise level (L1, L2, etc.) is also not a suitable learning indicator, as it can lead to producing quite blurry outputs while minimizing the function. %

\subsubsection{Perceptual losses}
We aim at removing shadows while preserving the non-shadowed areas as unaltered as possible.

Therefore, inspired by recent literature in photo-enhancement~\cite{ignatov2017dslr}, style transfer~\cite{gatys2016image} and perceptual super resolution~\cite{Johnson2016Perceptual}, we form a perceptual ensemble loss for color, style, and content, respectively. The parameters used were empirically chosen in relation to the amplitude of each loss on a subset of the train data: $\alpha_1 = 1$, $\alpha_2 = 0.1$ and $\alpha_3 =10000$.

\vspace{-5mm}
\begin{equation}
    L_{perceptual} = \alpha_{1} \cdot L_{color} + \alpha_{2} \cdot L_{content} + \alpha_{3} \cdot L_{style},
    \label{eq:ploss}
\end{equation}

\subsubsection{Color loss} 
The introduction of a color loss can be explained, firstly, by the need to capture and preserve color information in the image. Under ideal settings, this could be done by imposing a pixel-level loss (\eg, L1, L2, MSE). However, we consider that the color is a lower frequency component than the textural information of the image (our eyes are less sensitive to color than to intensity changes) and the pixel-level observations are generally noisy (\eg, pixel-wise inconsistencies in ISTD image pairs).
To this end, inspired by~\cite{Stoutz_2018_ECCV_Workshops}, we perform a Gaussian filter over the real and fake image, and compute the Mean Squared Error (MSE),

\vspace{-2mm}
\begin{equation}
    L_{color} = MSE(I_{smoothed}^1, I_{smoothed}^2)
    \label{eq:colloss}
\end{equation}

\subsubsection{Content loss}
Building on the assumption that an image with shadows and that one without shadows should have similar content in terms of semantic relevant regions, the $L_{content}$ is defined as
\vspace{-4mm}
\begin{equation}
    L_{content} = \frac{1}{N_l}\sum_{i=1}^{N_l}MSE(C_{I^1}^i, C_{I^2}^i),
    \label{eq:conloss}
\end{equation}
where $C^i$ is the feature vector representation extracted in the $i$-th target layer of the ImageNet pretrained VGG-16 network~\cite{simonyan2014very}, for each input image $I^{n}$. 

\subsubsection{Style loss} 
$L_{style}$, is defined as
\vspace{-2mm}
\begin{equation}
    L_{style} = \frac{1}{N_l}\sum_{i=1}^{N_l}MSE(H_{I^1}^i, H_{I^2}^i)
    \label{eq:stloss}
\end{equation}
\vspace{-2mm}
\begin{equation}
    {H_I^l}_{i,j} = \sum_{k=1}^{D} {C_I^l}_{i, k}{C_I^l}_{k,j}
\end{equation}
where the Gram matrix $H_I^l$ of the feature vector extracted by every $i$-th layer of the VGG-16 net. The Gram matrix $H^i_I$ defines a style for the feature set extracted by the $i$-th layer of the VGG-16 net, using as input the $I$ image. By minimizing the mean square error difference between the styles computed for feature sets at different levels of complexity, the results produced will be characterized by better perceptual properties. 

\begin{figure}[t]
\centering
\resizebox{\linewidth}{!}
{
\begin{tabular}{c|c}
\includegraphics[width=8cm]{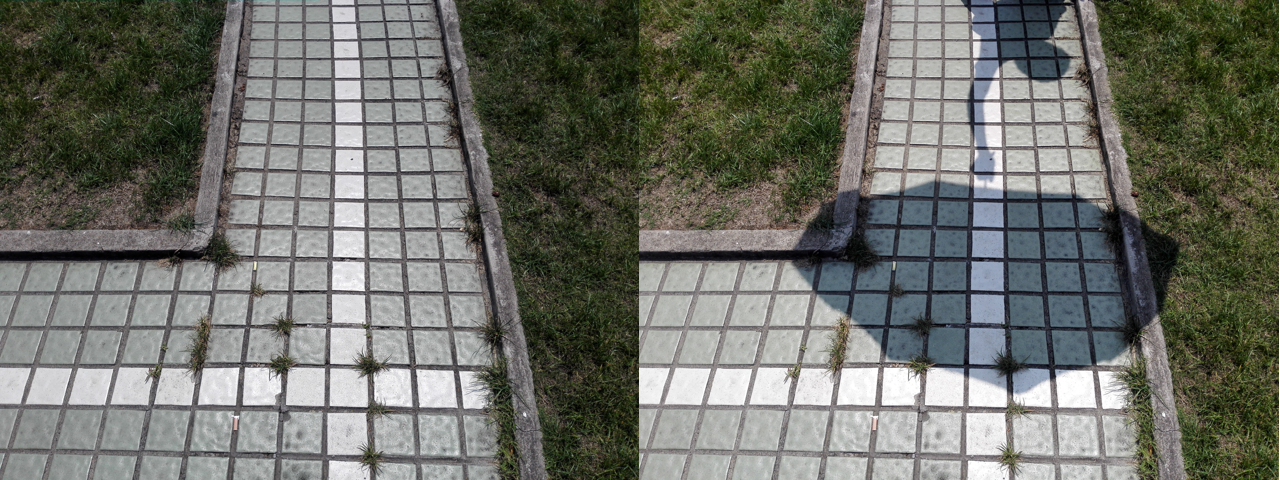}     & \includegraphics[width=8cm]{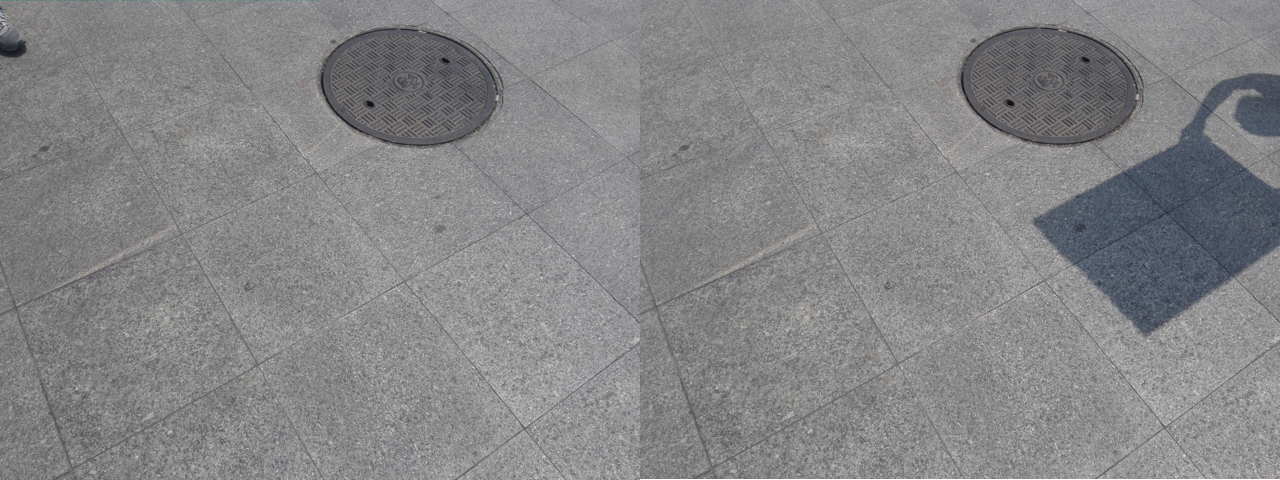} \\
a) illumination changes & b) semantic changes (check corner)\\
\end{tabular}
}
\caption{Examples of ISTD paired images that are not perfectly aligned and consistent.}
\label{fig:ISTD_differences}
\end{figure}

\subsubsection{Adversarial Losses}

The formulation of the problem using adversarial learning implies the introduction of two new components, $D_f$ and $D_s$. The main idea behind the learning procedure is that, for each domain, the discriminator will distinguish between the synthetic and the real results, forcing the counterpart generator to produce a better output in terms of semantic content and image properties.

As the discriminators are characteristic to the shadow domain $X$ and the shadow free domain $Y$, the adversarial losses are defined, for the synthetic results produced in the forward step ($\hat{I_f}$,$\hat{I_s}$), as stated in \eref{eq: gans} and \ref{eq: ganf}. The image pair $(I_s, I_f)$ is the ground truth shadow-shadow free pair used as input, and $(\hat{I_s}^{*}, \hat{I_f}^{*})$ is a pair of randomly sampled synthetic results.

\begin{equation}
    \begin{split}
        L_{GAN}^s (I_s, I_f) &= \frac{1}{2}(MSE(J, D_s(\hat{I_s}, I_s))\\
        &+ MSE(O, D_s(\hat{I_f}^{*}, I_s))), \ \forall \hat{I_f}^{*} \notin X
    \end{split}
    \label{eq: gans}
\end{equation}
\vspace{-4mm}
\begin{equation}
    \begin{split}
        L_{GAN}^f (I_s, I_f) &= \frac{1}{2}(MSE(J, D_f(\hat{I_f}, I_f))\\
        &+ MSE(O, D_f(\hat{I_s}^{*}, I_f))), \ \forall \hat{I_s}^{*} \notin Y
    \end{split}
       \label{eq: ganf}
\end{equation}

The standard output for the discriminators for the positive and negative examples used in training was defined as $J$ and $O$, as a consequence of the usage of the \emph{patchGAN} concept. So, $J$ and $O$ are defined as the all one matrix, and all zero matrix, respectively, with a size equal to the size of the perceptive field of the discriminator.

\subsubsection{Self-Supervised Shadow Loss}
\label{sec:s3}
The last and core ingredient of our system is the self-supervised shadow loss (S3 loss). In order to generate an image with shadows, we couple the shadow insertion generator to receive both a binary mask in addition to the RGB image. Our rationale comes from realizing that shadow addition can be at any random shape, scale, size, and position, then tackling this problem in an unconditional way is ill-posed. Additionally, by guiding the shadow insertion using a conditional mask, we can use a randomly inserted shadow mask into a deshadowed image in order to perform a cycle consistency loss to recover the mask.

Considering $u$ the shadow free image and $v$ the shadow image, then the generated images have the form of $\hat{u} = G_f(v)$ and $\hat{v} = G_s(u, m)$, where $m$ is the shadow mask between the images $u$ and $v$\footnote{We circumvent the problem of creating realistic random masks by using the mask on the counterpart forward step}. Similarly, we compute the reconstructed cycle-consistency images as $u_r = G_f(\hat{u})$ and $v_r = G_s(\hat{u}, \hat{m^f})$, where $\hat{m^f}=Bin(\hat{u}- v)$ is the synthetic shadow mask computed as the binarization of the difference between the synthetic shadow free image and the true shadow image. The shadow mask loss is defined as the L1 distance between the masks produced by the intermediate results produced along with the cycles.

The synthetic mask is a very important detail of the proposed model because is implied in two of the constraints we enforced over the model trained. For the fully supervised training, where the ground truth shadow mask is provided in the training/validation splits, one of the constraints is the invariance property of the shadow mask along the forward step of the cycle, in the shadow removal procedure. The second constraint is the invariance of the mask along the reconstruction step, both in shadow removal and addition, that can be applied both in the paired and unpaired settings. 

For the model trained in an unsupervised manner,  the used mask in the shadow addition step is sampled from a memory buffer, containing the masks produced in the shadow removal procedure, from the forward step\footnote{It is important to note that the synthetic masks are computed on the part of the cycle that the real mask does not exist.}. %will not be provided the real mask}.
As the generator uses this result in the reconstruction step, the quality of this result is crucial for a qualitative reconstruction. Even if a good mask will be the result of a realistic transformation, the usage of this proxy loss will enable faster learning, and also, the learning of more realistic mappings.

\subsubsection{Total Loss}
\label{sec:total_loss}
As the images that are to be fed to the model in the unpaired setting do not represent the same scene, the loss function has to be carefully chosen such that the equilibrium point can be reached, and so, the learning procedure will converge to the required solution. 
As we defined the GAN loss in \eref{eq: gans} and \ref{eq: ganf}, we are using replay buffers such that the discriminators are less likely to rely on the simple difference between frames representing similar contents. Thus, the degree of control is weaker, as we are sampling from intermediary results to feed the negative samples in the binary classification problem. So, additional information can be exploited by using different complexity features extracted by a pretrained neural network.

Moreover, another crucial goal is controlling, as much as possible, the quality of the intermediary results, by the way of minimizing the distance between the topology information localizing the hallucinated area in the forward pass, to the one observed in the reconstruction pass of the training step. So, we are exploiting the observation that, under convergence conditions, both shadow removal and shadow addition procedures are inverse to each other, as we stated shadow removal as a bijectivity in the problem formulation. 

\begin{equation}
\begin{split}
    L_{gen} (u,v) &= \gamma_1 \cdot (L_{GAN}^f(\hat{u}, u) + L_{GAN}^s( \hat{v}, v) \\
    &+ L_{GAN}^f(u_r, u) + L_{GAN}^s(v_r, v))  \\
    &+ \gamma_2 \cdot (L_{content}(u, \hat{v}) + L_{content}(v, \hat{u})) \\
    &+ \gamma_3 \cdot (L_{pix}(u, u_r) + L_{pix}(v,v_r)) \\  % + \beta_1 L_{pix}(u, \hat{u})) \\
    &+ \gamma_4 \cdot (L_{perceptual}(u, u_r) + L_{perceptual}(v, v_r)) \\
    &+ \gamma_5 \cdot (L_{mask}(\hat{m^f}, m_r^f) + L_{mask}(\hat{m^s}, m_r^s)\\
    &+ \beta_2 L_{mask}(\hat{m}^{*}, \hat{m^f})    ,
\end{split}
\label{eq: totalloss}
\end{equation}
%\vspace{-2mm}.

So, we choose the total loss for the unpaired case as a linear combination of the losses described, where $\gamma$ and $\beta$ parameters control the contribution of each loss. Note that each component is easily extracted from~\fref{fig:overview}.
As the shadow position and shape are invariant under the transformations implied, by adding this term in the training objective, the transformations will be naturally plausible. As the datasets are not characterized by a high variation in terms of shadow regions shapes and positions, the model will benefit from adding a loss term such that the mask produced in the reconstruction procedure is similar to the sampled mask $\hat{m}^{*}$

\subsubsection{Total Loss for Paired Data}
\label{sec:total_loss_paired}
As our method can be extended for paired datasets, in~\eref{eq:totalloss_paired} we show the modifications to the loss functions in this scenario.
\vspace{-2mm}
\begin{equation}
\begin{split}
    L_{gen} (u,v) &= \gamma_1 \cdot (L_{GAN}^f(\hat{u}, u) + L_{GAN}^s( \hat{v}, v) \\
    &+ L_{GAN}^f(u_r, u) + L_{GAN}^s(v_r, v))\\
    &+ \gamma_2 \cdot (L_{content}(u, \hat{v}) + L_{content}(v, \hat{u})) \\
    &+ \gamma_3 \cdot (L_{pix}(u, u_r) + L_{pix}(v,v_r) + \beta_1 L_{pix}(u, \hat{u})) \\
    &+ \gamma_4 \cdot (L_{perceptual}(u, u_r) + L_{perceptual}(v, v_r)) \\
    &+ \gamma_5 \cdot (L_{mask}(\hat{m^f}, m_r^f) + L_{mask}(\hat{m^s}, m_r^s)\\
    &+ \beta_2 L_{mask}(m, \hat{m^f}))
\end{split}
\label{eq:totalloss_paired}
\end{equation}
When training using paired data, a constraint can be used to speed-up the convergence process, by adding the term $ L_{pix}(u, \hat{u})$ as the L1 pixel-wise loss between the input shadow free image and the shadow free image generated in the forward step of the cycle. The $\beta_2$ parameter is needed to force the model to create a suitable transformation in the forward step of the cycle, as the reconstruction process will be using this intermediate representation of the shadow free image. As the mask shape and position should be the same, the shadow masks would not differ along the cycle, so $L_{mask}$, the L1 distance between the two shadow masks, was introduced. The weights used in the linear combination of GAN, L1, perceptual, or mask losses were determined with respect to the magnitude of each term, and their speed of decrease. Exact values were provided in \tref{tab:parameters}, for both the paired and the unpaired settings.

\begin{table}[t]
\begin{center}
\caption{Parameters of the total loss function (\eref{eq: totalloss} and \ref{eq:totalloss_paired}) defined for our training framework.}
\resizebox{\linewidth}{!}{
     \begin{tabular}{c|cccccccc}
         \hline
        Setting & $\gamma_1$ & $\gamma_2$ & $\gamma_3$ & $\gamma_4$ & $\gamma_5$ & $\beta_1$ & $\beta_2$ \\ [0.5ex] 
         \hline
        \hline 
         Unpaired training  & 250 & 10 & 100 & 30 & 60 & 0 & 100
         \\ 
         Paired training & 250 & 20 & 60 & 50 & 60 & 10 & 100 \\ 
         \hline
     \end{tabular}
 }
 \label{tab:parameters}
 \end{center}
\end{table}

\subsection{Implementation details}
\label{ssc:implementation_details}

\begin{table*}[ht]
\centering
\caption{General details about the architecture of the generators. \emph{LR} is LeakyReLU(0.2), \emph{R} is ReLU, and \emph{TH} is the hyperbolic tangent activation function.}
\resizebox{\linewidth}{!}{
 \begin{tabular}{| c | c |c | c | c| c | c | c | c|c |c | c | c| c | c | c | c|} 
 \hline
    Layer & 1 & 2 & 3 & 4 & 5 & 6 & 7 & 8 & 9 & 10 & 11 & 12 & 13 & 14 & 15 & 16 \\  
 \hline
    Channels in & 3/4 & 64 & 128 & 256 & 512 & 512 & 512 & 512 & 512 & 1024 & 1024 & 1024 & 1024 & 512 & 256 & 64 \\
 \hline
    Channels out & 64 & 128 & 256 & 512 & 512 & 512 & 512 & 512 & 512 & 512 & 512 & 512 & 256 & 128 & 64 & 3 \\
 \hline
    Operation & $o_1$ & $o_1$ & $o_1$ & $o_1$ & $o_1$ & $o_1$ & $o_1$ & $o_1$ & $o_2$ & $o_2$ & $o_2$ & $o_2$ & $o_2$ & $o_2$ & $o_2$ & $o_3 $\\
 \hline
    Normalization & 0 & 1 & 1 & 1 & 1 & 1 & 1 & 0 & 1 & 1 & 1 & 1 & 1 & 1 & 1 & 0 \\
 \hline
    Activation & LR & LR & LR & LR & LR & LR & LR & LR & R & R & R & R & R & R & R & TH \\
 \hline
    Dropout & 0 & 0 & 0 & 0.5 & 0.5 & 0.5 & 0.5 & 0.5 & 0.5 & 0.5 & 0.5 & 0.5 & 0 & 0 & 0 & 0 \\
  \hline  
 \end{tabular}
 }
 \label{tab:tgen}
\end{table*}

\paragraph{\textbf{Generator.}}
Details of the \textit{generator} implementation are provided in~\tref{tab:tgen}, where the operation $o_1$ is a convolutional operation with kernel size 4, stride 2 and padding 1 and $o_2$ is its upsampling counterpart that uses a transposed convolution. Skip-connections were added between the downsampling blocks and the upsampling counterparts. The $o_3$ operation, present in the last layer of the architecture, is a convolution with kernel size 4 and padding 1, preceded by an upsampling with scale factor 2 and zero padding 1 in the top and the bottom size of the feature tensor. The result is then passed into the $\tanh$ activation, obtaining the corresponding pixel in the produced image.
\vspace{-4mm}
\paragraph{\textbf{Discriminator.}}
The \textit{discriminator} consists of four convolutional blocks, each having the convolution operator with $k=4$, instance normalization and LeakyReLU as activation ($a=0.2$). The final output size of the discriminator will be the size of the patch described as the ``perceptive field'' of the model. The depth of the initial input tensor can be explained by the fact that the discriminator will receive as input a pair of images, each of them with three channels, as they are RGB images.
\vspace{-5mm}
\paragraph{\textbf{Initialization.}}
As the initialization, the weights in both the discriminators and the generators were drawn from a Gaussian distribution with 0 mean and 0.2 variance.
%%%%%%%%%%%%%%%%%%%%%%%%%%%%%%%%%%%%%%%%%%%%%%%%%%%%%%%%%%%%

%%%%%%%%%%%%%%%RESULTS%%%%%%%%%%%%%%%%%%%%%%%%%%%%%%%%%%%%%%

\section{Experimental Results}
\label{sec:experimental_results}

\subsection{Setup}
\label{ssc:setup}

\paragraph{\textbf{Datasets.}}
We validate our system over \emph{ISTD}~\cite{ISTDwang2018STCGAN}, and \emph{USR}~\cite{USRhu2019mask} datasets. On the one hand, the ISTD dataset contains paired data for shadow and shadow-free images. Given the illumination inconsistency problem in this dataset, Le~\etal~\cite{Le_2019_ICCV} proposed a compensation method, creating thus the \emph{ISTD+} dataset. On the other hand, the USR dataset is a collection of unpaired shadow and shadow free images used for unsupervised tasks.

For unpaired training and testing on the ISTD dataset, the shadow and shadow free images were randomly sampled. The random mask inserted in each iteration comes from a buffer bank of real masks.
\vspace{-5mm}
\paragraph{\textbf{Experimental Framework.}}

We train our system during 100 epochs, learning rate $0.005$ with $\lambda$-decay scheduling after the first 40 epochs. 
We use Adam optimizer with  $\beta = (0.9, 0.999)$. 

For both the paired and unpaired settings, the masks were computed as a binarization of the difference between the shadow free image and the shadow image, by a thresholding procedure using the median value of the difference.

\paragraph{\textbf{Evaluation measures.}}

For the quantitative evaluation of our method, we use the Root Mean Square Error (RMSE) and Peak Signal to Noise Ratio (PSNR) between the output deshadowed image and the reference/ground truth image. We compute these pixel-wise fidelity measures in both RGB and Lab color space, respectively.

It is well established that RMSE and PSNR do not correlate well with perceptual quality, so complementary to the fidelity measures, we also employ LPIPS~\cite{zhang2018perceptual} score in order to assess the photo-realism of the produced deshadowed images with respect to the ground truth.

\paragraph{\textbf{Compared methods.}}
We directly compare our proposed solution to two other methods capable to learn from unpaired data: \emph{CycleGAN}~\cite{CycleGAN2017} and \emph{Mask Shadow GAN}~\cite{USRhu2019mask}. Moreover, in order to compare with prior systems in paired learning, we report qualitative and quantitative results for the following methods:
\emph{DSC}~\cite{hu2019direction}, \emph{ST-CGAN}~\cite{ISTDwang2018STCGAN} and \emph{DeShadowNet}~\cite{SRDDESHADOW}.

\subsection{Ablation Study}
% \AR{Is there going to be an ablation study? I saw that was a main issue from ECCV reviews}
In~\tref{tab:abb-results} we report our results on ISTD for different settings. Each configuration was trained for 100 epochs, on ISTD dataset. When switching from a learning procedure based on both fidelity and perceptual losses to an only-fidelity loss based objective ($\gamma_4 = 0$) the results improve in fidelity and lack in perceptual terms. The removal of the mask loss ($\gamma_5 = 0$) produces similar results in terms of both perceptual and fidelity measures, but the standard deviation over the LPIPS score is higher due to a more pronounced difficulty of the model to deal with more complex textures.

The forward pixel-wise loss ($\beta_1=0$) is very important in order to produce realistic results in the forward step of the cycle, even though the latent representation learnt ($\hat{u},\hat{v}$), for both shadow and shadow free domains, produce the best results in terms of reconstruction error (either fidelity loss or perceptual score). As a better mask is a consequence of a better reconstruction, the dropping of the forward mask ($\beta_2=0$), produce results characterized by similar pixel-wise properties, but lacking in perceptual terms.

\begin{table}[t]
    \centering
        \caption{The impact of various loss function parameters on the performance.}    
    \label{tab:abb-results}
    \resizebox{\linewidth}{!}
    {
    \begin{tabular}{l||cc|cc|cc||}
    &\multicolumn{2}{c|}{LPIPS$\downarrow$}&\multicolumn{2}{c|}{RMSE$\downarrow$}&\multicolumn{2}{c||}{PSNR$\uparrow$} \\
    Method &avg&stddev&RGB&Lab&RGB&Lab\\
    \hline
    Paired setting & \textbf{0.031} & 0.025 & 15.05 & 4.18 & 26.88 & 37.91  \\
    %\hline
    Paired setting ($\gamma_2=0$) & 0.032 & 0.029 & 14.75 & 4.15 & 27.1 & 38.05 \\
    %\hline
    Paired setting ($\gamma_4=0$) & 0.033 & 0.022 & 14.07 & 4.01 & 27.39 &  38.31\\
    %\hline
    Paired setting ($\gamma_5=0$) & 0.031 & 0.027 & \textbf{13.80} & \textbf{3.90} & \textbf{27.70} & \textbf{38.58} \\
    %\hline
    %\hline
    Paired setting ($\beta_2=0$) & 0.035 & \textbf{0.019} & 14.70 & 4.11 & 27.26 & 38.2 \\
    \hline
    Paired setting\tablefootnote{This configuration was not considered, due to the random forward mapping effect} ($\beta_1=0$) & 0.021 & 0.046 & 5.96 & 2.61 & 34.27 & 41.67 \\
     \end{tabular}
    }

\end{table}

\begin{table}[h]
    \caption{Ablative results in terms of pixel-wise loss (RMSE and PSNR, both on Lab space) and perceptual quality loss (LPIPS) for different settings of the loss function.}
    \label{tab: abb_results_unpaired}
    \begin{center}
         \begin{tabular}{c || c c c ||} 
             %\hline
             Setting description & RMSE$\downarrow$ & PSNR$\uparrow$ & LPIPS$\downarrow$ \\
             \hline
             default set of parameters & 5.73 & 33.40 & \textbf{0.045}\\ 
             \hline
             default and $\gamma_2 = 0$ & 5.34 & 34.35 & 0.082\\
             %\hline
             default and $\gamma_3 = 10$ & 5.98 & 32.87 & 0.104 \\
             %\hline
             default and $\gamma_4 = 0$ & \textbf{3.71} & \textbf{37.31}  & 0.056\\
             %\hline
             default and $\gamma_5 = 0$ & 4.42 & 36.56  & 0.074\\ [1ex] 
            % \hline
        \end{tabular}
    \end{center}
\end{table}

In \tref{tab: abb_results_unpaired}, we investigated the behaviour of our model under different configurations of the unpaired setting. A trade-off between improving in terms of fidelity score vs. the perceptual properties can be observed, concluding that both the mask loss and the perceptual loss yield better results in terms of perceptual score.

\subsection{Quantitative results}
To quantitatively evaluate the performance of our shadow removal solution, we adhere to the ISTD and USR benchmarks~\cite{USRhu2019mask,ISTDwang2018STCGAN} and report the results in~\tref{tab:results}. For all the reported results, we used our models trained on the training partition of the ISTD dataset, for 100 epochs for both the paired and the unpaired settings. For the unpaired setting, the shadow and shadow free training images were sampled without replacement. 
The USR dataset provides a collection of shadow free images and two splits of shadow images, for training and validation, which are not representing the same scene as the shadow-free images. The same sampling procedure was deployed for the USR dataset.

As shown in~\tref{tab:results} our models largely improve the state-of-the-art in both fidelity (RMSE, PSNR) and perceptual measures (LPIPS) on both benchmarks.

\begin{table}[t]
        \caption{Comparison with state-of-the-art methods on ISTD and USR datasets.}    
    \label{tab:results}
    \centering
    \resizebox{1.05\linewidth}{!}
    {
      \hspace{-0.75cm}
    \begin{tabular}{l||cc|cc|cc||cc|cc|cc}
    &\multicolumn{6}{c||}{ISTD test images}&\multicolumn{6}{c}{USR test images}\\
    &\multicolumn{2}{c|}{LPIPS$\downarrow$}&\multicolumn{2}{c|}{RMSE$\downarrow$}&\multicolumn{2}{c||}{PSNR$\uparrow$}&\multicolumn{2}{c|}{LPIPS$\downarrow$}&\multicolumn{2}{c|}{RMSE$\downarrow$}&\multicolumn{2}{c}{PSNR$\uparrow$}\\
Method &avg&stddev&RGB&Lab&RGB&Lab&avg&stddev&RGB&Lab&RGB&Lab\\
    \hline\hline
    \multicolumn{13}{c}{Unpaired data training}\\
    \hline
    MaskShadowGAN\cite{USRhu2019mask}&0.25 & 0.09 & 28.34 & 7.32 & 19.78 & 31.65  & 0.31 & 0.11  & 27.53 & 7.06 & 19.97 & 31.76\\
    CycleGAN~\cite{CycleGAN2017} & 0.118 & 0.07 & 25.4 & 6.95 & 20.59  & 31.83 & 0.147 & 0.07 & 30.04 & 9.66 & 19.04 & 29.06 \\
    \textbf{ours (unpaired)}& \textbf{0.041} & \textbf{0.033} & \textbf{7.58} & \textbf{5.12} & \textbf{31.18} & \textbf{34.45} & \textbf{0.009} & \textbf{0.004} & \textbf{5.70} & \textbf{2.21} & \textbf{33.26} & \textbf{41.06}\\

    \hline
        \multicolumn{13}{c}{Paired data training}\\
    \hline
    DeShadowNet~\cite{SRDDESHADOW} & 0.080 & 0.055 & 31.96 & 7.98 & 19.30 & 31.27 & - & - & - & - & - & - \\    
    DSC~\cite{hu2019direction} &0.202 & 0.087 & 23.36 & 6.03 & 21.85 & 33.63 & - & - & - & - & - & - \\
    ST-CGAN~\cite{ISTDwang2018STCGAN} & 0.067 & 0.043 & 22.11 & 5.93 & 22.66 & 34.05 & - & - & - & - & - & - \\
    \textbf{ours (paired)} & \textbf{0.031} & \textbf{0.025} & \textbf{15.05} & \textbf{4.18} & \textbf{26.88} & \textbf{37.90} & - & - & - & - & - & - \\
    \hline
    \end{tabular}
    }
\end{table}

\begin{table}[]
    \caption{Lab color space results for both shadow and shadow-free pixels on ISTD\cite{ISTDwang2018STCGAN} and ISTD+\cite{Le_2019_ICCV} datasets.}
    \vspace{-3mm}
    \centering
    \resizebox{\linewidth}{!}
    {
    \begin{tabular}{l|cc||cc|cc|cc}
    %\hline
\multicolumn{3}{c||}{Setup} & \multicolumn{2}{c|}{All} & \multicolumn{2}{c|}{Shadow} & \multicolumn{2}{c}{Shadow free} \\
Method&Train&Test & RMSE & PSNR & RMSE & PSNR & RMSE & PSNR \\
\hline

\hline
\textbf{ours (unpaired)} & ISTD & ISTD & 5.12 & 34.45 & 6.98 & 32.65 & 4.94 & 34.71\\
\textbf{ours (paired)}& ISTD & ISTD & 4.18 & 37.90 & 4.63 & 36.87 & 4.07 & 38.22\\
\hline
\textbf{ours (paired)}& ISTD+& ISTD+ & \textbf{3.04} & \textbf{41.10} & \textbf{4.15} & \textbf{38.16} & \textbf{2.77} & \textbf{42.05} \\

\cite{Le_2019_ICCV} (paired)& ISTD+
& ISTD+ & 3.8 & n/a & 7.4 & n/a & 3.1 & n/a\\
\hline
\end{tabular}
}
    \label{tab:ISTD+}
\end{table}

\begin{figure*}[ht]
    \centering
    \resizebox{1.05\linewidth}{!}
    {
    \hspace{-1.5cm}
    \begin{tabular}{c||cc||cccc||c}
    \Large input & \multicolumn{2}{c||}{\Large Unpaired data training} & \multicolumn{4}{c||}{\Large Paired data training} & \Large ground truth\\ 
    shadow   & 
Mask Shadow GAN & 
\Large \textbf{ours (unpaired)} & 
\Large \textbf{ours (paired)} &
DeShadowNet  & 
DSC & 
ST-CGAN & 
shadow free\\
\includegraphics[width=0.25\linewidth, height=0.25\linewidth]{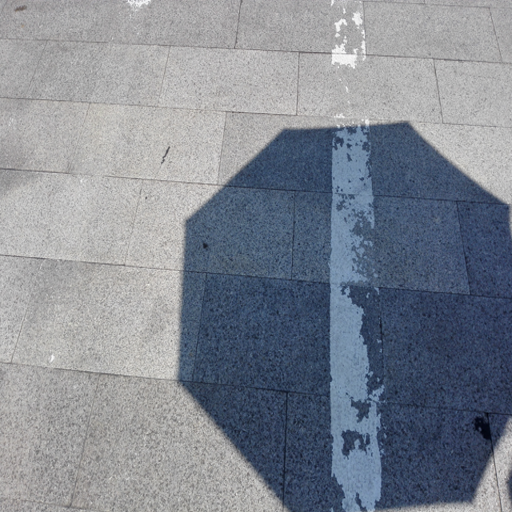} &
\includegraphics[width=0.25\linewidth, height=0.25\linewidth]{ 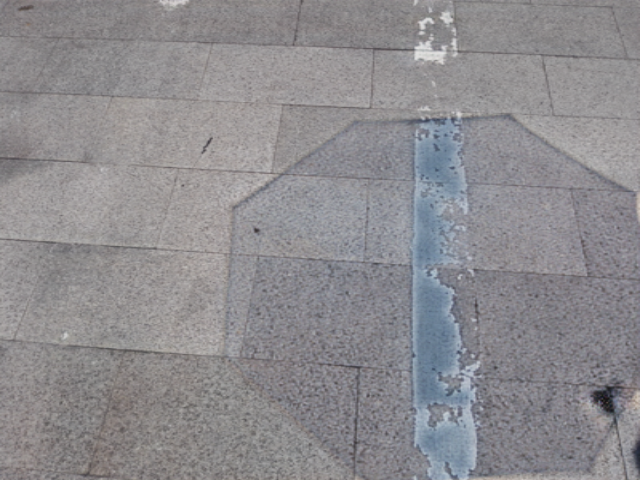}&
\includegraphics[width=0.25\linewidth]{ 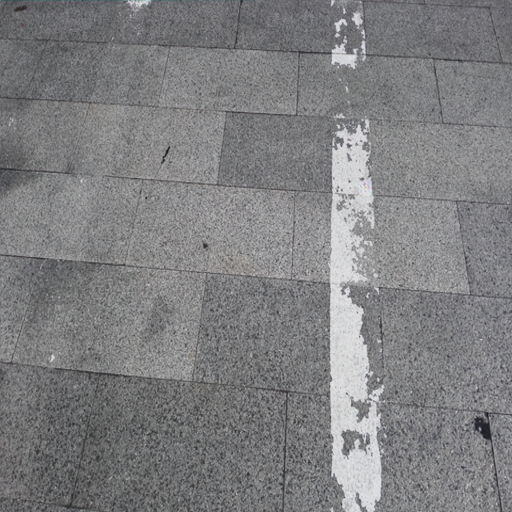}&
\includegraphics[width=0.25\linewidth]{ 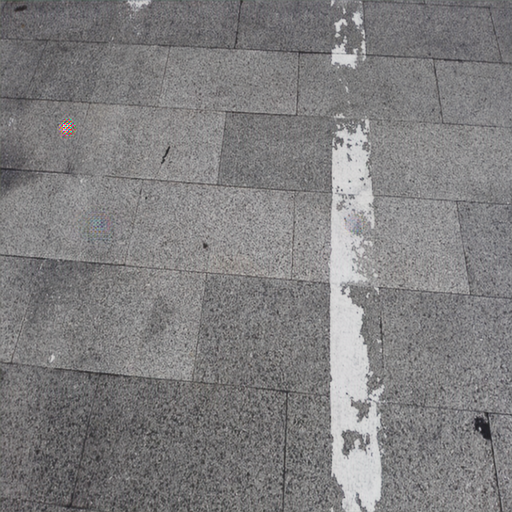}&
\includegraphics[width=0.25\linewidth, height=0.25\linewidth]{ 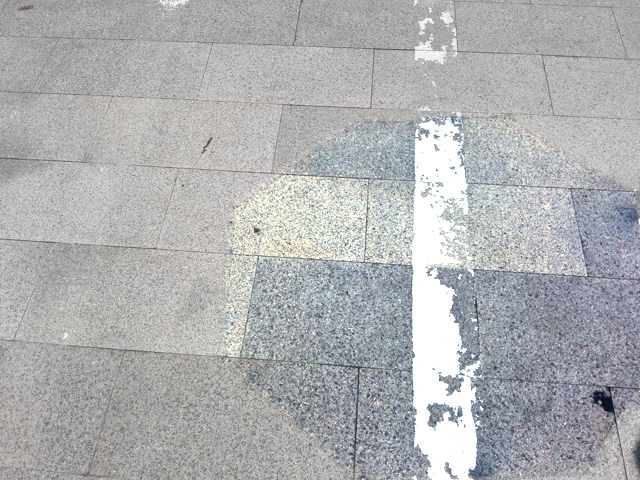} &
\includegraphics[width=0.25\linewidth, height=0.25\linewidth]{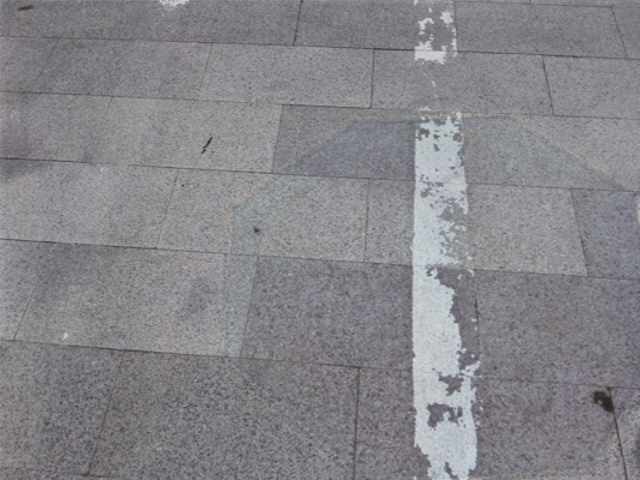} &
\includegraphics[width=0.25\linewidth]{ 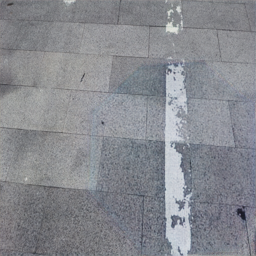} &
\includegraphics[width=0.25\linewidth]{ 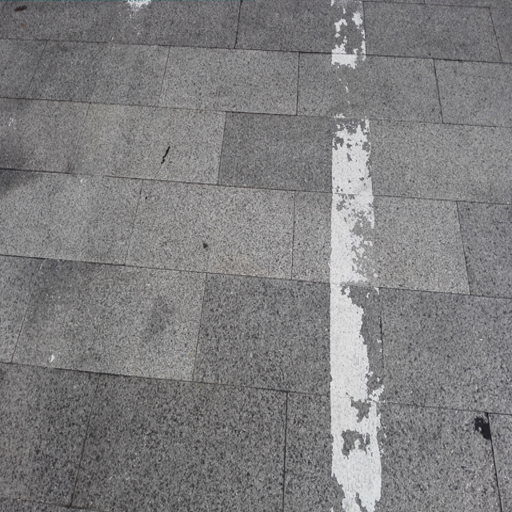} \\
\includegraphics[width=0.25\linewidth, height=0.25\linewidth]{ 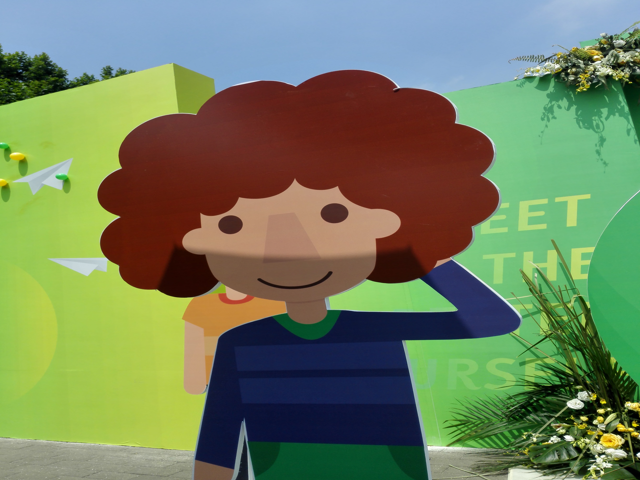} &
\includegraphics[width=0.25\linewidth, height=0.25\linewidth]{ 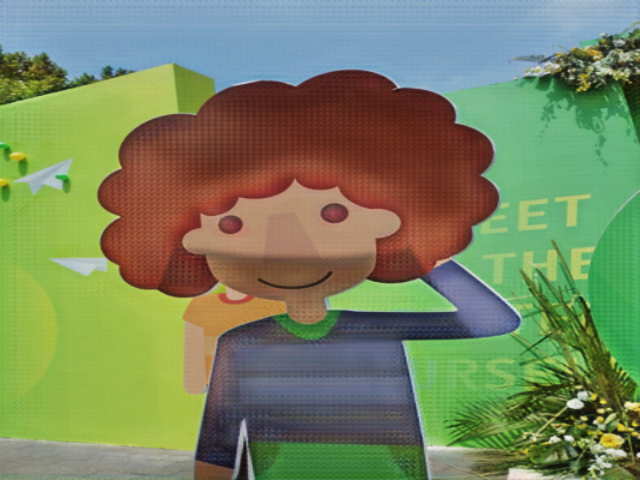}&
\includegraphics[width=0.25\linewidth]{ 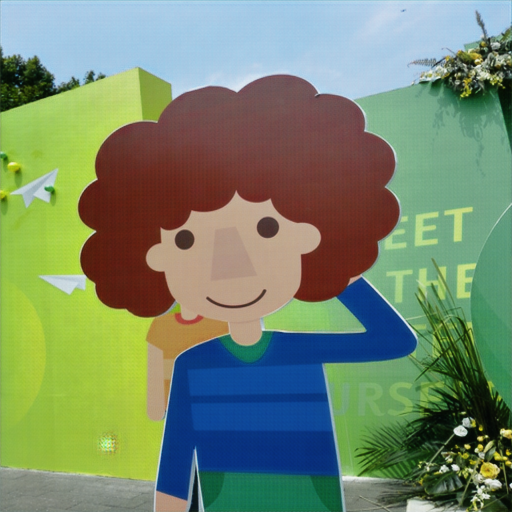}&
\includegraphics[width=0.25\linewidth]{ 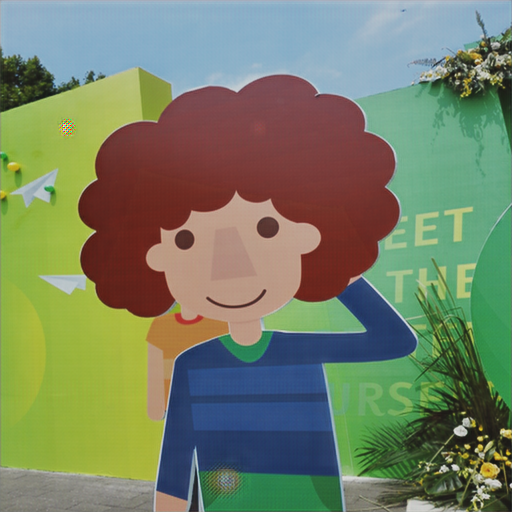} &
\includegraphics[width=0.25\linewidth, height=0.25\linewidth]{ 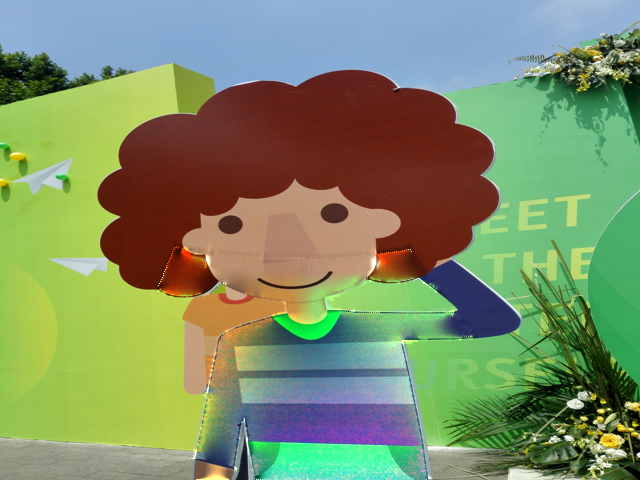} &
\includegraphics[width=0.25\linewidth, height=0.25\linewidth]{ 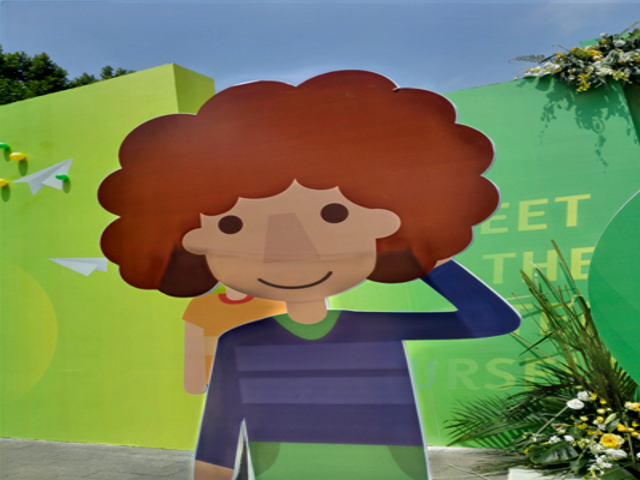} &
\includegraphics[width=0.25\linewidth]{ 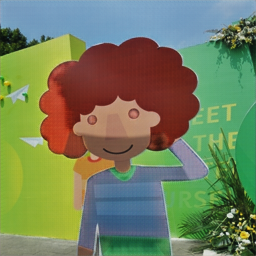} &
\includegraphics[width=0.25\linewidth]{ 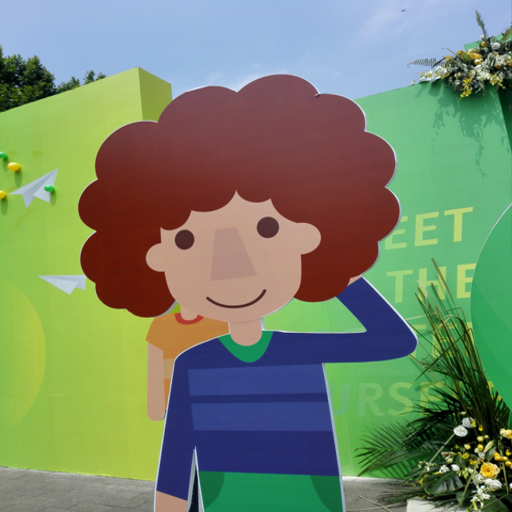} \\

\includegraphics[width=0.25\linewidth, height=0.25\linewidth]{ 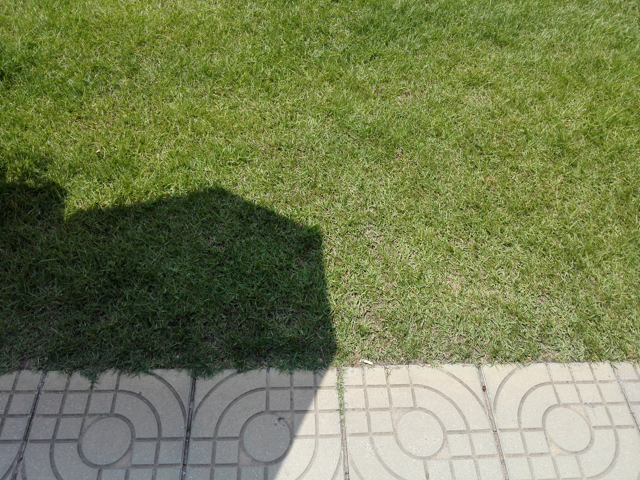} &
\includegraphics[width=0.25\linewidth, height=0.25\linewidth]{ 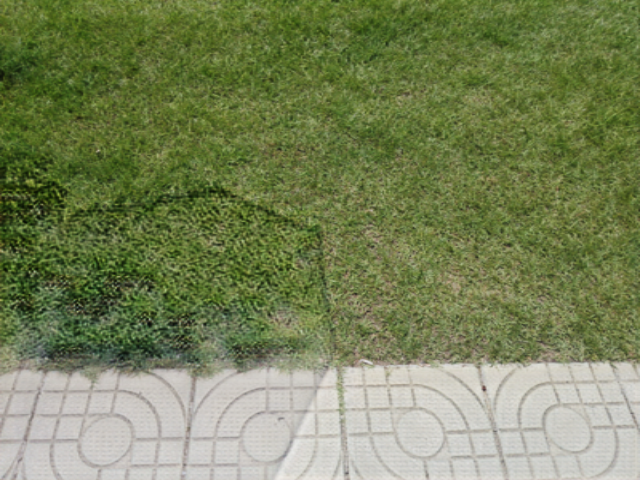}&
\includegraphics[width=0.25\linewidth]{ 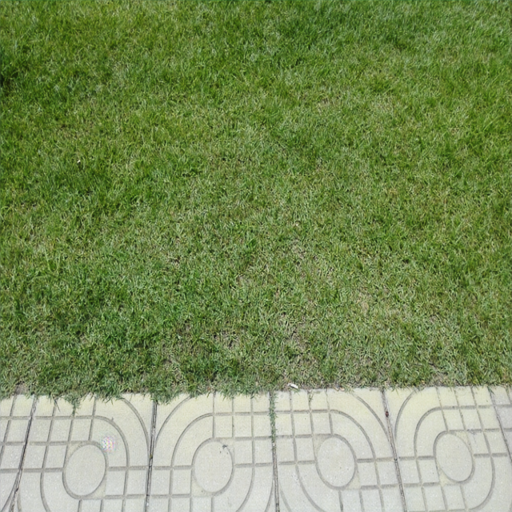}&
\includegraphics[width=0.25\linewidth]{ 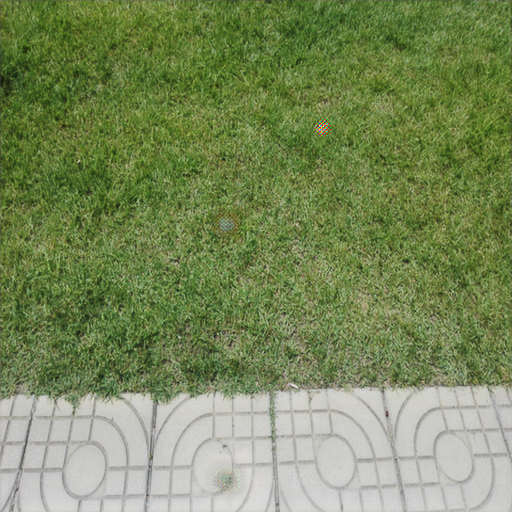}&
\includegraphics[width=0.25\linewidth, height=0.25\linewidth]{ 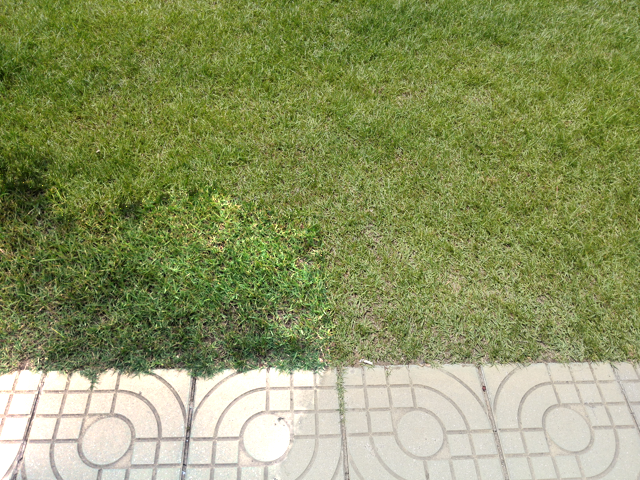} &
\includegraphics[width=0.25\linewidth, height=0.25\linewidth]{ 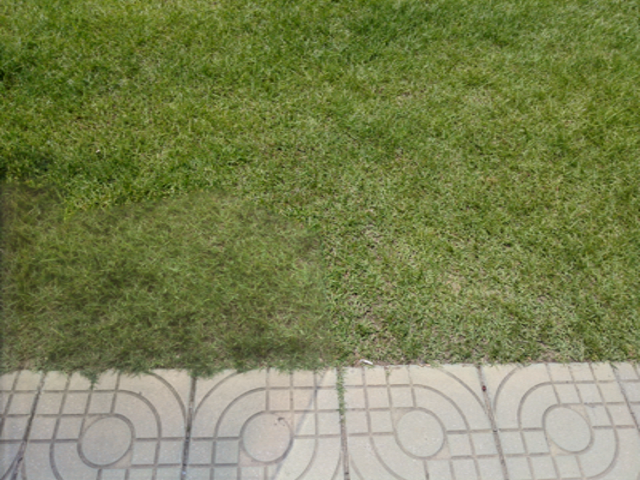} &
\includegraphics[width=0.25\linewidth]{ 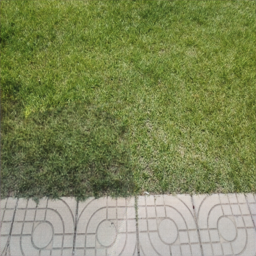} &
\includegraphics[width=0.25\linewidth]{ 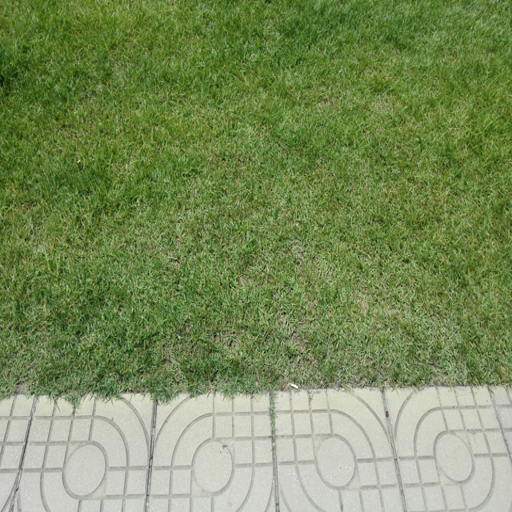} \\
\includegraphics[width=0.25\linewidth, height=0.25\linewidth]{ 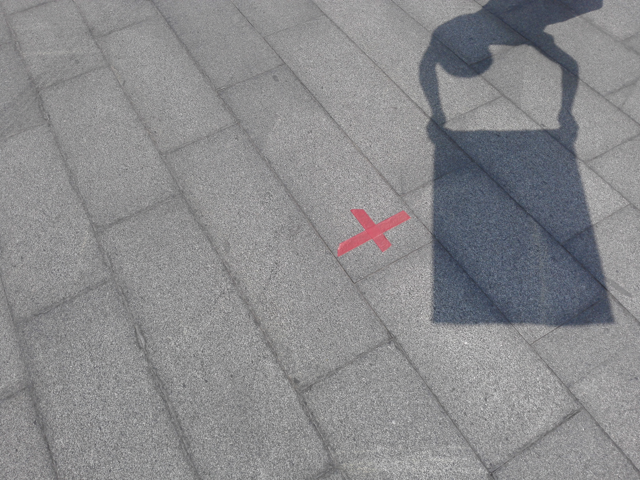} &
\includegraphics[width=0.25\linewidth, height=0.25\linewidth]{ 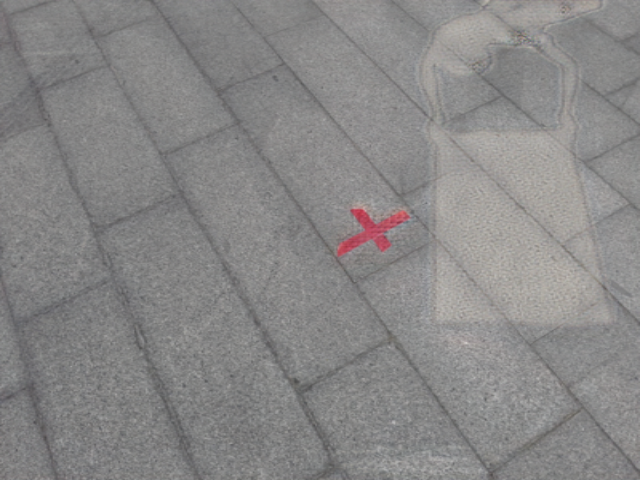}&
\includegraphics[width=0.25\linewidth]{ 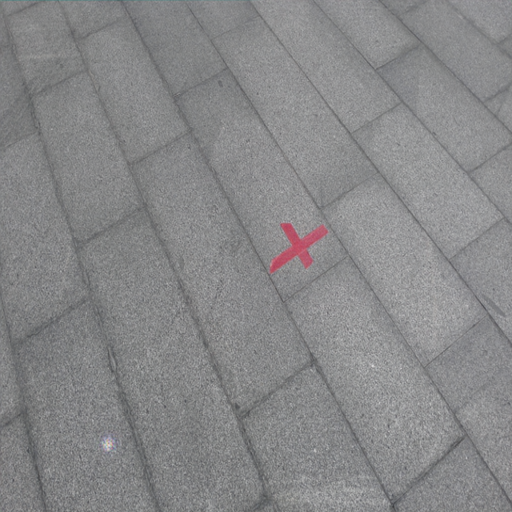}&
\includegraphics[width=0.25\linewidth]{ 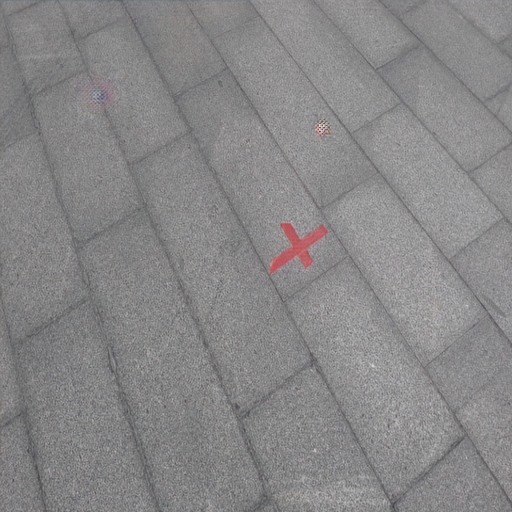}&
\includegraphics[width=0.25\linewidth, height=0.25\linewidth]{ 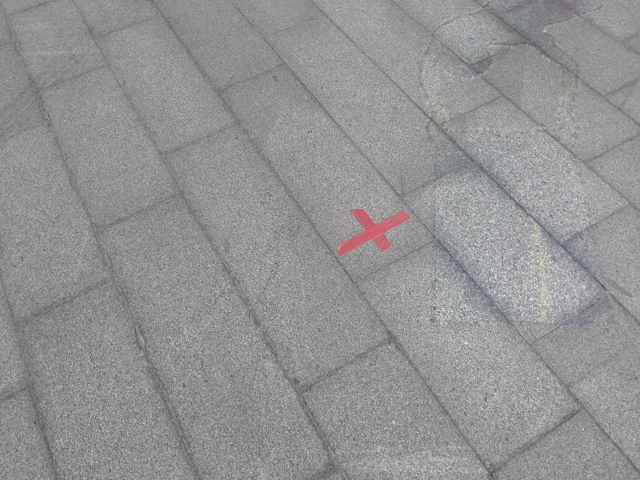} &
\includegraphics[width=0.25\linewidth, height=0.25\linewidth]{ 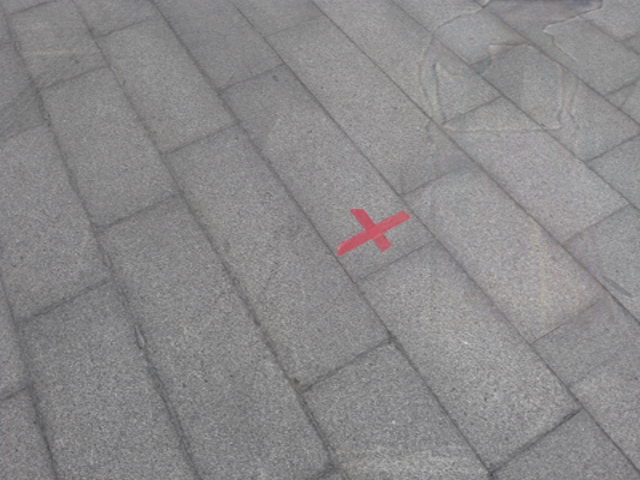} &
\includegraphics[width=0.25\linewidth]{ 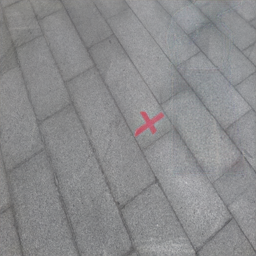} &
\includegraphics[width=0.25\linewidth]{ 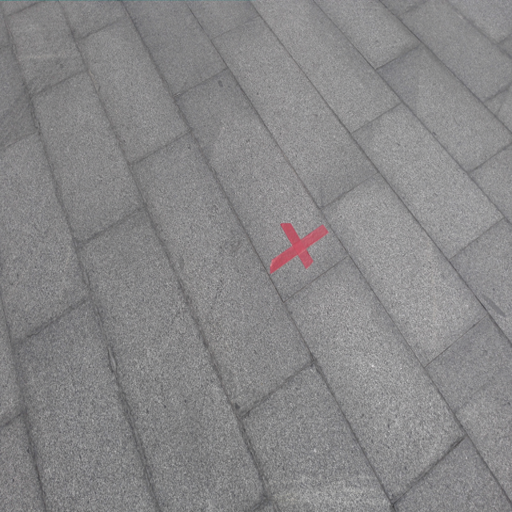} \\

    \end{tabular}
    }
    \caption{Visual results for the proposed solution and comparison with state-of-the-art learned methods. Best zoom in on screen for better details. }
    \label{fig:visual_comparison1}
\end{figure*}

\subsection{Qualitative results}
\label{ssc: qual_results}

\fref{fig:visual_comparison1} shows visual results obtained on randomly picked ISTD test images. 
We note that the results achieved by our solutions are the closest to the reference shadow free images, while the other methods generally produce strong artifacts.

Our results clearly improve over the unpaired state-of-the-art Mask Shadow GAN, producing more appealing and artifact-free images. On the paired counterpart, our method completely removes the shadow while the related methods produce visible traces.

\begin{figure}[t]
    \centering
    \resizebox{0.9\linewidth}{!}
    {
    \begin{tabular}{c|cc|c|cc}
          
    input & ours (unpaired) & ours (paired) &
    
    input & ours (unpaired) & ours (paired) \\
\includegraphics[width=0.25\linewidth, height=0.25\linewidth]{ 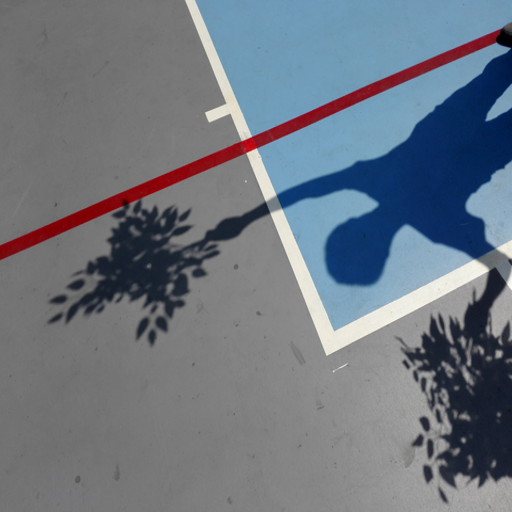} & 
\includegraphics[width=0.25\linewidth, height=0.25\linewidth]{ 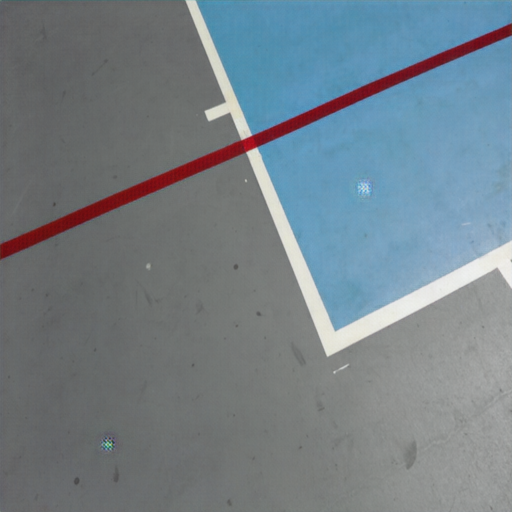} & 
\includegraphics[width=0.25\linewidth, height=0.25\linewidth]{ 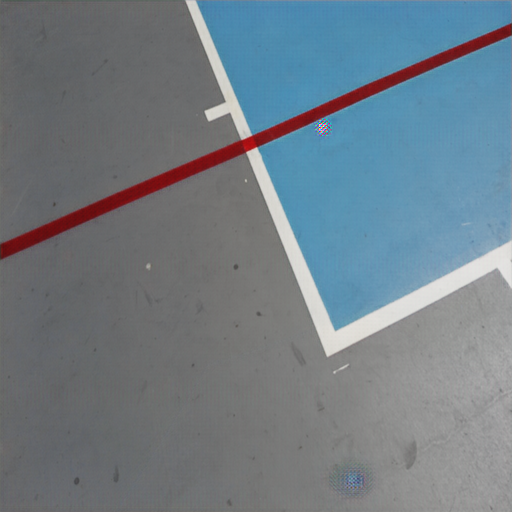} & 
\includegraphics[width=0.25\linewidth, height=0.25\linewidth]{ 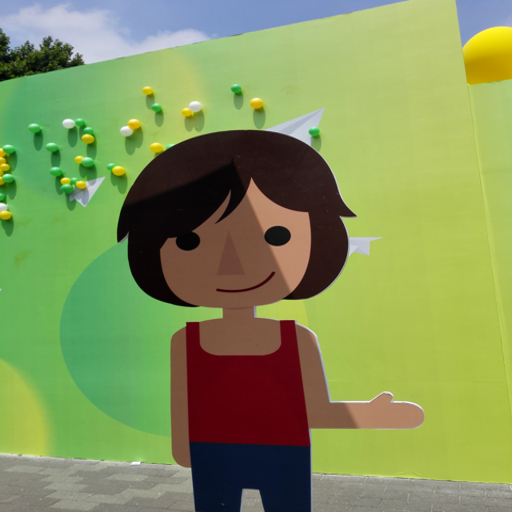} & 
\includegraphics[width=0.25\linewidth, height=0.25\linewidth]{ 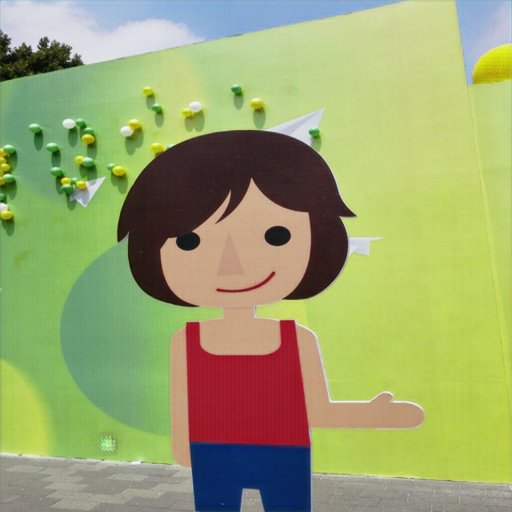} &
\includegraphics[width=0.25\linewidth, height=0.25\linewidth]{ 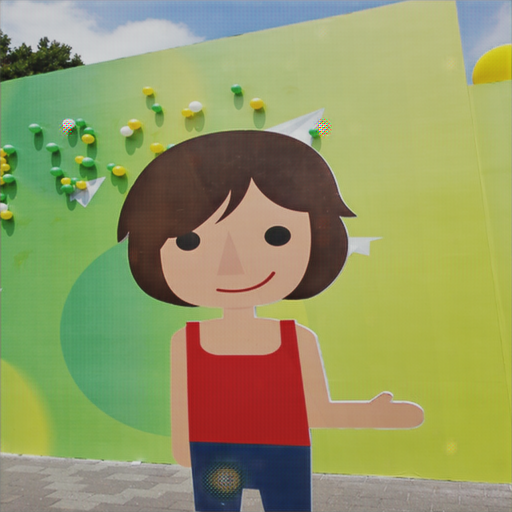}
\\
ground truth          &error heatmap & error heatmap &
ground truth          &error heatmap & error heatmap\\
\includegraphics[width=0.25\linewidth, height=0.25\linewidth]{ 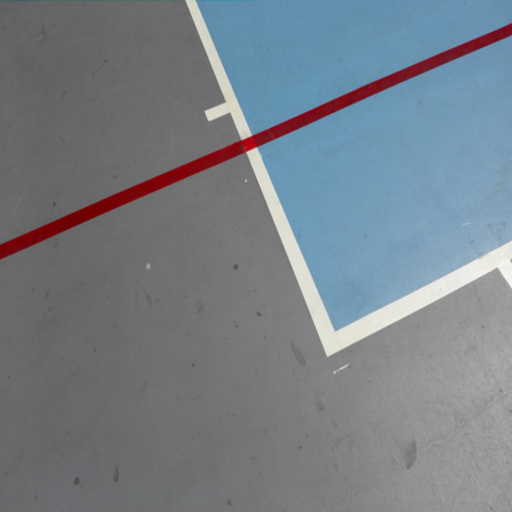} 

& 
\includegraphics[width=0.25\linewidth, height=0.25\linewidth]{ 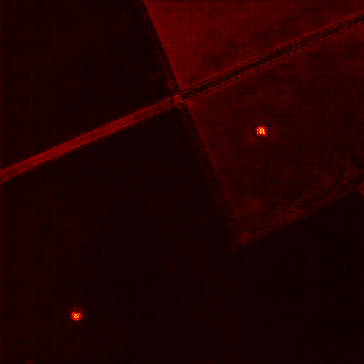} 
& 
\includegraphics[width=0.25\linewidth, height=0.25\linewidth]{ 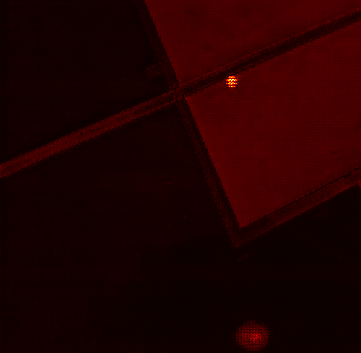} 
& 
\includegraphics[width=0.25\linewidth, height=0.25\linewidth]{ 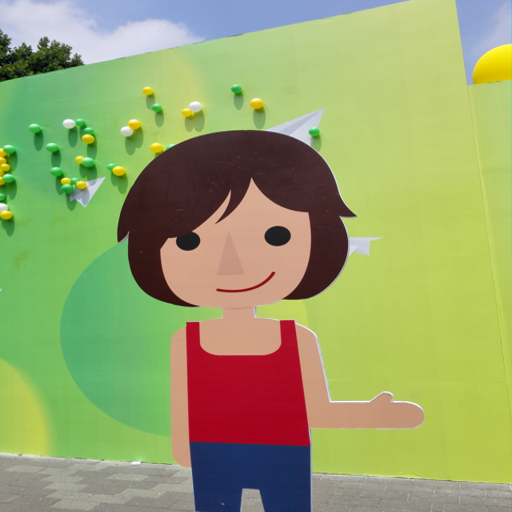} 
& 
\includegraphics[width=0.25\linewidth, height=0.25\linewidth]{ 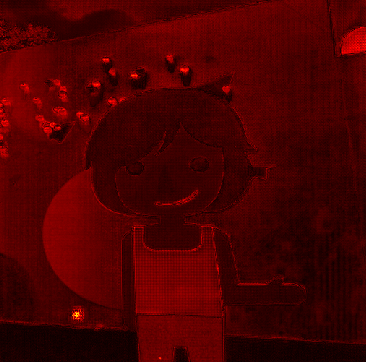} 
& 
\includegraphics[width=0.25\linewidth, height=0.25\linewidth]{ 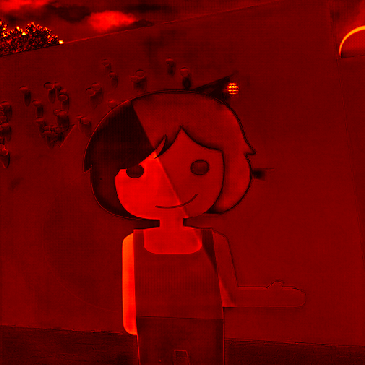} 
\\    
\end{tabular}
    }
    \caption{Visual results for the proposed solutions trained with unpaired and paired, and corresponding error heatmaps. Best zoom in on screen for better details.}
    \label{fig:heatmaps}
\end{figure}
\subsection{Discussion}
For both paired and unpaired settings, our system produces the best perceptual metrics (lower LPIPS) and the best pixel-wise error metrics (PSNR and RMSE) with respect to state-of-the-art methods by large margins. 

\fref{fig:heatmaps} shows results and the square L2 norm of the residuals in the image space for our models. 
We observe that the paired version of the model has problems in recovering the unshadowed region on the neighborhoods characterized by sharp variations in terms of color and illumination variations. This could be due to the ISTD dataset used for training the model. ISTD has a limited shadow formation diversity in its pairs. Therefore, the model provides poorer results on images representing much more complex scenes. 

Furthermore, as we show in~\fref{fig:ISTD_differences}, the semantic differences and the differences in illumination are expected to induce a certain degree of uncertainty when using largely weighted L1 loss terms between ground truth images and the synthetically generated images in the same domain. Therefore, as it can be seen in~\fref{fig:heatmaps}, the error is not concentrated in the shadow affected area, but, in steep variations in terms of texture, and some peaks in error can be observed in that area. When training in an unpaired manner, by simply dropping this loss term we can overcome this issue, improving our results on the ISTD dataset, compared to the paired setting.

The unpaired version also benefits from both sampling processes deployed, \textit{i.e.}, for the shadow mask (using a mask buffer) and the negative examples for discriminator training. Since the sample sets are dynamically generated from synthetic data, the variation of the provided examples is expected to be higher. Therefore, the generalization ability of the model increases (as it can be observed in~\tref{tab:results}) producing better results in terms of both pixel-wise loss and perceptual metrics.

This behaviour can be explained  by  the model benefiting  from the variety of random localization/shape combinations characterising the shadowed region.

Additionally, the higher generalization ability of the discriminators provides another degree of control such that a  robust learning procedure converges to a realistic mapping from the shadow domain $X$ to the shadow free domain $Y$.

Although the degree of control is weak under the unpaired setting, the exploitation of both deep features, and the proxy loss defined for the transformed region provides sufficient information for the learnt mapping to be realistic.

%%%%%%%%%%%%%%%%%%%%%%%%%%%%%%%%%%%%%%%%%%%%%%%%%%%%%%%%%%%%

%%%%%%%%%%%%%%%%%%CONCLUSIONS%%%%%%%%%%%%%%%%%%%%%%%%%%%%%%%

\section{Conclusions}
\label{sec:conclusions}
In this work we proposed a novel unsupervised single image shadow removal solution. We rely on self-supervision and jointly learn shadow removal from and shadow addition to images. As our experimental results show on ISTD and USR datasets, we set a new state-of-the-art in single image shadow removal, by largely outperforming prior works in both fidelity (RMSE, PSNR) and perceptual quality (LPIPS) for both paired and unpaired settings.

%%%%%%%%%%%%%%%%%%%%%%%%%%%%%%%%%%%%%%%%%%%%%%%%%%%%%%%%%%%%

\section*{Acknowledgments}
This work was partly supported by the ETH Zürich Fund (OK), a Huawei project, an Amazon AWS grant, and an Nvidia hardware grant.

{\small
\bibliographystyle{ieee_fullname}
\bibliography{sssr_wacv}
}

\end{document}